%% file: main.tex
\documentclass[10pt,twocolumn,letterpaper]{article}

\usepackage{cvpr}              


\definecolor{cvprblue}{rgb}{0.21,0.49,0.74}
\usepackage[pagebackref,breaklinks,colorlinks,allcolors=cvprblue]{hyperref}

\def\paperID{1290} 
\def\confName{CVPR}
\def\confYear{2025}

\title{APT: Adaptive Personalized Training for Diffusion Models with Limited Data}

\author{JungWoo Chae$^{1}$\thanks{Equal contribution} \quad 
Jiyoon Kim$^{1}$\footnotemark[1] \quad
JaeWoong Choi$^{1}$ \quad
Kyungyul Kim$^{1}$ \quad
Sangheum Hwang$^{2}$\thanks{Corresponding author}\\[6pt]
{\normalsize
\begin{tabular}{@{}c@{\quad}c@{}}
      $^1$LG CNS AI Research & 
      $^2$\begin{tabular}[t]{@{}l@{}}
        Department of Data Science,
        Seoul National University of Science and Technology
      \end{tabular}
    \end{tabular}
}\\
{\tt\small{\{cjwoolgcns, jiyoonkim, jaewoong\_choi, kyungyul.kim \}@lgcns.com} }\\
{\tt\small  shwang@ds.seoultech.ac.kr}
}

\input{macro.tex}
\usepackage{dblfloatfix}
\usepackage{capt-of}
\usepackage{afterpage}
\usepackage{float}
\usepackage{cuted}
\usepackage{booktabs}
\usepackage{tabularx}
\usepackage{layouts}
\usepackage{balance} 

\begin{document}

\maketitle

\begin{strip}
\vspace{-40pt}
\centering
\includegraphics[width=\linewidth]{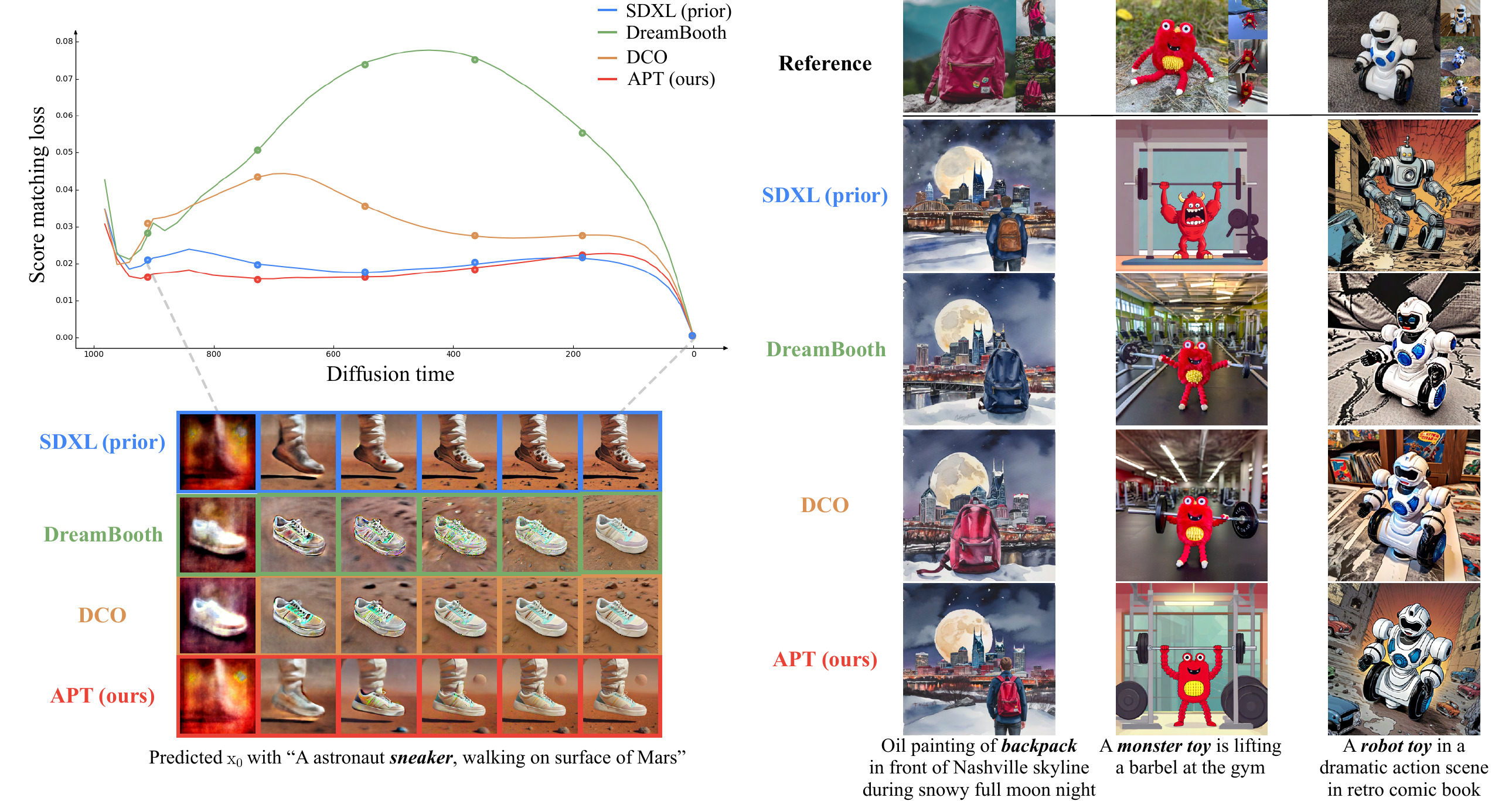}
\captionof{figure}{Given a few reference images, APT personalizes diffusion models with less overfitting: (Left) By comparing diffusion trajectories using the score matching loss~\cite{cfg++}, we observe that our method maintains the original denoising path. The predicted $\mathbf{x}_0$ images from APT closely resemble SDXL (prior) during early steps, preserving the overall layout and scene context. (Right) APT effectively incorporates contextual elements from the prior, such as generating a backpack with a person without explicitly mentioning ``person'' and preserves stylistic elements like comic book aesthetics. In contrast, other methods either focus excessively on reference images or fail to maintain the prior's style. This demonstrates that APT successfully maintains the pretrained model's capabilities for text alignment and stylization.}

\label{fig:teaser}
\end{strip}

\input{sec/0_abstract}

\input{sec/1_introduction}

\input{sec/2_rel_work}

\input{sec/3_method}
\input{sec/4_experiment}

\input{sec/5_conclusion}

\clearpage

\subsection*{Acknowledgement} This research was supported by Seoul National University of Science and Technology (2024-0200).

{
    \small
    \bibliographystyle{ieeenat_fullname}
    \bibliography{main}
}
\input{sec/11_suppl}


\end{document}

%% file: macro.tex
\newcommand{\ours}{\textbf{APT}}

%% file: sec/0_abstract.tex
\begin{abstract}

Personalizing diffusion models using limited data presents significant challenges, including overfitting, loss of prior knowledge, and degradation of text alignment. Overfitting leads to shifts in the noise prediction distribution, disrupting the denoising trajectory and causing the model to lose semantic coherence. In this paper, we propose \textbf{Adaptive Personalized Training (\ours)}, a novel framework that mitigates overfitting by employing adaptive training strategies and regularizing the model's internal representations during fine-tuning. APT consists of three key components: (1) \textbf{Adaptive Training Adjustment}, which introduces an overfitting indicator to detect the degree of overfitting at each time step bin and applies adaptive data augmentation and adaptive loss weighting based on this indicator; (2) \textbf{Representation Stabilization}, which regularizes the mean and variance of intermediate feature maps to prevent excessive shifts in noise prediction; and (3) \textbf{Attention Alignment for Prior Knowledge Preservation}, which aligns the cross-attention maps of the fine-tuned model with those of the pretrained model to maintain prior knowledge and semantic coherence. Through extensive experiments, we demonstrate that \ours \ effectively mitigates overfitting, preserves prior knowledge, and outperforms existing methods in generating high-quality, diverse images with limited reference data.

\end{abstract}


%% file: sec/1_introduction.tex
\section{Introduction}
\label{sec:introduction}

The advent of diffusion models has significantly advanced the field of generative modeling, enabling the synthesis of diverse and high-quality images~\cite{ddpm, ddim, ldm}. Personalization techniques, such as DreamBooth~\cite{dreambooth} and Textual Inversion~\cite{ti}, have further enhanced these models by enabling subject-driven generation tailored to specific user needs. Such advancements have broad applications, from artistic content creation to specialized data augmentation in machine learning tasks~\cite{jedi, disendiff}.
However, personalizing diffusion models using limited data presents significant challenges. One critical issue is overfitting, which causes excessive shifts in the noise prediction distribution, disrupting the denoising trajectory of the pretrained diffusion model~\cite{dreambooth, dco}. These shifts lead to the loss of prior knowledge, degradation of text alignment, and a reduced ability of the model to generalize to unseen prompts. 

Overfitting may cause the model to memorize spatial layouts, resulting in generated images with overly similar compositions, or to over-memorize textures, leading to poor stylization and a lack of diversity in response to different prompts. Moreover, overfitting can lead to the loss of prior knowledge, causing the model to generate images that do not accurately reflect the desired concept or context. For example, when generating an image with \emph{``a photo of a backpack''}, pretrained diffusion models may naturally include a person carrying the backpack, leveraging prior knowledge about common contexts. However, after fine-tuning with limited data, the model may lose this prior knowledge, resulting in images of only the backpack without a person. This loss of prior knowledge is accompanied by changes in the cross-attention maps, which further degrade the quality and coherence of the generated images.

Existing methods \cite{dco,face2diff,svdiff,disendiff,attndreambooth} have addressed these challenges through various regularization techniques and novel fine-tuning approaches. Techniques that constrain attention using masks~\cite{break_a_scene} often require additional annotations and may not align effectively with the soft attention distributions of the model. Furthermore, prior preservation techniques that incorporate additional data~\cite{dreambooth} fine-tune the model by combining subject and auxiliary images but often suffer from overfitting. This can disrupt the original denoising trajectory, resulting in overfitting to auxiliary datasets and reduced generalization in generation quality~\cite{dco}.

In this work, we propose \textbf{Adaptive Personalized Training (\ours)}, a novel framework that addresses these challenges by mitigating overfitting with adaptive training strategies, regularizing the internal representations of the model during fine-tuning, and preserving prior knowledge. Specifically, our method consists of three key components:




\begin{enumerate}
\item \textbf{Adaptive Training Adjustment}: We introduce an overfitting indicator to detect the degree of overfitting and apply adaptive data augmentation and loss weighting based on this indicator. This approach addresses the varying influence of diffusion model parameters across different time steps due to the beta scheduling, effectively mitigating overfitting and adjusting the training dynamics.

\item \textbf{Representation Stabilization}: We regularize the significant shifts in the noise prediction $\boldsymbol{\epsilon}$ caused by overfitting by constraining the mean and variance of the intermediate feature maps. This helps preserve the statistical properties of the representations of the pretrained model.

\item \textbf{Attention Alignment for Prior Knowledge Preservation}: To maintain the prior knowledge in the text embeddings, we propose regularizing the cross-attention maps. By aligning the attention distributions of the fine-tuned model with those of the pretrained model, we ensure that the model retains semantic coherence.
\end{enumerate}

\noindent Our contributions can be summarized as follows:

\begin{itemize}



\item We introduce \textbf{Adaptive Personalized Training (\ours)}, a novel method that addresses overfitting and the loss of prior knowledge in diffusion model personalization with limited data. APT incorporates adaptive training adjustments, representation stabilization, and attention alignment to mitigate overfitting and preserve prior knowledge during fine-tuning.

\item Through extensive experiments, we demonstrate that APT outperforms existing methods in preserving the text alignment ability and prior knowledge of pretrained models, while generating high-quality and diverse personalized images.

\end{itemize}

Our method provides a cohesive solution that addresses both the varying influence of model parameters across time steps and the internal representation shifts that arise during fine-tuning with limited data. By mitigating overfitting and preserving prior knowledge, we enable the model to generalize better to unseen prompts while accurately capturing the desired concepts from the reference data.

%% file: sec/2_rel_work.tex
\section{Related Work}
\label{sec:related_work}



\paragraph{Text-to-Image Personalization}
Recent advances in diffusion models have enabled high-quality image synthesis through large-scale datasets and advanced architectures~\cite{ldm, imagen, dalle2, dalle3, parti, glide, palette}, with techniques like classifier-free guidance~\cite{cfg, cfgplus, cfgrethinking} enhancing text alignment.

Personalization of text-to-image models adapts pretrained models to represent new concepts based on user-provided images. Key methods include DreamBooth~\cite{dreambooth}, which fine-tunes the entire model for high fidelity, and Textual Inversion~\cite{ti}, which optimizes textual embeddings without altering model weights for efficiency.
Parameter-efficient fine-tuning methods, such as LoRA~\cite{lora}, Custom Diffision~\cite{customdiffusion}, and Svdiff~\cite{svdiff}, update only a small subset of parameters to reduce resource demands while maintaining quality. Recent advances like P+~\cite{p+} and NeTI~\cite{neti} expand textual conditioning spaces, enabling greater control and expressiveness without full model fine-tuning, achieving faster convergence and improved editability.

\vspace{-10pt}
\paragraph{Regularization in T2I Personalization}


Maintaining the prior knowledge of pretrained models during personalization is essential to prevent concept drift. Techniques like the prior preservation loss in DreamBooth~\cite{dreambooth} limit deviations from the original distribution but struggle with limited data, leading to inconsistencies and undesirable shifts~\cite{dco}.
Recent methods like DCO~\cite{dco} address this by directly regularizing the KL divergence, while Attention Regularization~\cite{attndreambooth} improves identity preservation through refined cross-attention maps. However, these methods primarily target specific components, such as cross-attention, and fail to fully preserve the pretrained model's diffusion trajectories, affecting text alignment and diversity.


To overcome these limitations, we propose a method that regularizes not only cross-attention but also self-attention, as well as intermediate representations such as U-Net outputs. By aligning these components along the diffusion process, our approach preserves the pretrained model's original capabilities while enabling accurate personalization.


\begin{figure*}[!htbp]
  \centering
  \includegraphics[width=0.85\linewidth]{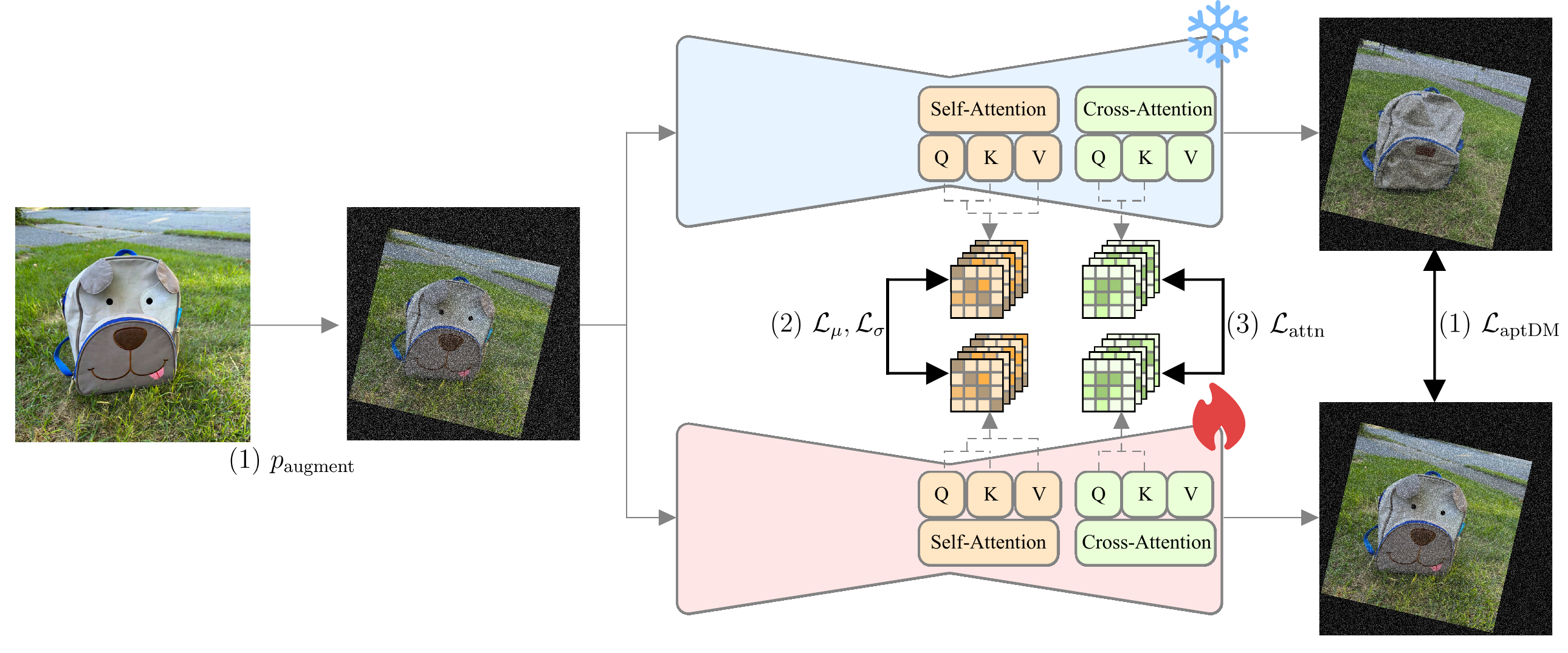}
  \caption{\textbf{Overview of Our Proposed Method (\ours).} Illustration of the three key components: (1) Adaptive Training Adjustment with adaptive data augmentation ($p_{\text{augment}}$) and loss weighting ($\mathcal{L}_{\text{aptDM}}$) to mitigate overfitting; (2) Representation Stabilization through regularizing intermediate feature maps to stabilize the noise trajectory ($\mathcal{L}_{\mu},\mathcal{L}_{\sigma}$); (3) Attention Alignment to preserve prior knowledge by regularizing the cross-attention maps ($\mathcal{L}_{\text{attn}}$).}
  \label{fig:overview}
\end{figure*}

\vspace{-10pt}
\paragraph{Adaptive Data Augmentation in Generative Models}


Overfitting in generative models trained on limited data is a critical challenge. StyleGAN ADA~\cite{stylegan_ada} addresses this in GANs by applying augmentations adaptively based on the degree of overfitting, stabilizing training without modifying loss functions or network architectures. Improved Consistency Reguarlization~\cite{bcr} similarly enhances GANs by enforcing consistency on the discriminator.



While these methods target GANs, adaptive augmentation to mitigate overfitting is also relevant for diffusion models. In our work, we introduce an adaptive augmentation strategy based on the proposed overfitting indicator, dynamically adjusting augmentation strength to prevent overfitting in personalization.

%% file: sec/3_method.tex
\section{Method}
\label{sec:method}


Personalizing diffusion models with limited reference data introduces significant challenges, such as overfitting, loss of prior knowledge, and degradation of text alignment \cite{dreambooth,dco,ti}. To address these issues, we propose \textbf{Adaptive Personalized Training (\ours)}, a method focused on mitigating overfitting through adaptive training strategies (Section~\ref{sec:ATA}), stabilizing the model's internal representations during fine-tuning (Section~\ref{sec:RS}), and preserving prior knowledge (Section~\ref{sec:AA}). An overview of our method is illustrated in Figure~\ref{fig:overview}.

\subsection{Adaptive Training Adjustment}
\label{sec:ATA}

Fine-tuning diffusion models on limited data can lead to overfitting, where the model excessively memorizes the training data. Due to the beta scheduling in diffusion models, the loss magnitude varies greatly across different time steps, affecting model updates differently at each step. This overfitting causes significant shifts in noise prediction $\boldsymbol{\epsilon}$, disrupting the denoising trajectory of the pretrained diffusion model, as observed in Figure~\ref{fig:teaser}. This disruption results in the degradation of text alignment and loss of prior knowledge.
Therefore, it is necessary to detect and mitigate overfitting by introducing an overfitting indicator and applying adaptive strategies based on it. By adjusting the training dynamics adaptively, we aim to mitigate overfitting and maintain the integrity of the denoising trajectory.

\vspace{-10pt}
\paragraph{Adaptive Overfitting Indicator}

We introduce an adaptive overfitting indicator $\gamma_t$ to quantify the degree of overfitting during fine-tuning:
\begin{equation}
\gamma_t = 1 - e^{- T \left( \text{EMA}_t\left[ \mathcal{L}_{\text{DM}}^{\phi} \right] - \text{EMA}_t\left[ \mathcal{L}_{\text{DM}}^{\theta} \right] \right)},
\label{eq:indicator}
\end{equation}
where $T$ is the total number of denoising steps, $\mathcal{L}_{\text{DM}}^{\phi}$ and $\mathcal{L}_{\text{DM}}^{\theta}$ are the denoising losses of the pretrained model $\phi$ and the fine-tuned model $\theta$, respectively. The $\text{EMA}_t$ denotes the exponential moving average computed at the specific time step bin $t$ to reduce fluctuations due to noise and data variance. This formulation ensures that $\gamma_t = 0$ when there is no overfitting and $\gamma_t \rightarrow 1$ when overfitting is maximal.
In practice, we divide the total diffusion steps into $B$ bins (e.g., $B = 10$ bins of 100 steps each for a total of 1000 steps). The overfitting indicator $\gamma_t$ is computed separately for each bin $t$, capturing the degree of overfitting at different noise levels.

\vspace{-10pt}
\paragraph{Adaptive Data Augmentation}

We use $\gamma_t$ as the data augmentation probability, clamping it within a predefined range:
\begin{equation}
p_{\text{augment}} = \texttt{clamp}\left( \gamma_{t}, 0, p_{\text{max}} \right),
\end{equation}
where $p_{\text{max}}$ is the maximum augmentation probability.  
As shown in Figure~\ref{fig:teaser}, the personalized model $\theta$ tends to memorize spatial configurations from early denoising steps, leading to positional overfitting. To disrupt this memorization, we apply affine transformations as data augmentation. By adjusting the probability of applying data augmentation based on $\gamma_t$, we aim to mitigate spatial overfitting.

\vspace{-10pt}
\paragraph{Adaptive Loss Weighting}

In addition to adaptive data augmentation, we adjust the loss weighting adaptively according to the degree of overfitting. We design a weighting scheme that scales the loss for each time step bin based on the degree of overfitting\footnote{The motivation for adaptive loss weighting is described in Supplementary Material B.3.}:
\begin{equation}
\mathcal{L}_{\text{aptDM}} = (1 - \gamma_t) \mathcal{L}_{\text{DM}},
\label{eq:loss_weighting}
\end{equation}
where $\gamma_t$ is the overfitting indicator for time step bin $t$, and $\mathcal{L}_{\text{DM}}$ is the denoising loss. By scaling the loss with $(1 - \gamma_t)$, we reduce its impact for time steps where overfitting is detected, effectively rebalancing the training dynamics and mitigating overfitting.

\subsection{Representation Stabilization}
\label{sec:RS}



To prevent the denoising trajectory of the fine-tuned model from deviating excessively from the original (i.e., the pretrained model's trajectory), it is necessary to regularize these shifts by stabilizing the intermediate representations. 


We apply regularization to the mean and variance of the intermediate feature maps of the model to preserve the statistical properties of the representations of the pretrained model. Let $\mathbf{h}_{\theta}^{(l)}$ and $\mathbf{h}_{\phi}^{(l)}$ denote the activations at layer $l$ for the fine-tuned model $\theta$ and the pretrained model $\phi$, respectively. We define the representation regularization losses as:
\begin{equation}
\mathcal{L}_{\mu} = \sum_{l}^{\text{layers}} \left| \mu\left(\mathbf{h}_{\theta}^{(l)}(x_t; c^*, t)\right) - \mu\left(\mathbf{h}_{\phi}^{(l)}(x_t; c, t)\right) \right|_2^2,
\label{eq:mean_reg}
\end{equation}
\begin{equation}
\mathcal{L}_{\sigma} = \sum_{l}^{\text{layers}} \left| \sigma\left(\mathbf{h}_{\theta}^{(l)}(x_t; c^*, t)\right) - \sigma\left(\mathbf{h}_{\phi}^{(l)}(x_t; c, t)\right) \right|_2^2,
\label{eq:var_reg}
\end{equation}
where $c^*$ is the conditioning information including the identifier (e.g., ``V*'') while $c$ is the conditioning information with the class token (e.g., ``dog''). $\mu(\cdot)$ and $\sigma(\cdot)$ compute the mean and standard deviation of activations, respectively. By regularizing these statistics, we limit excessive shifts in the distribution of the intermediate representations, preserving prior knowledge and improving text alignment.

\subsection{Attention Alignment for Prior Preservation}
\label{sec:AA}

Overfitting can lead to the loss of prior knowledge specified by the text embeddings, causing the model to generate images that do not accurately reflect the desired context. For example, when learning a concept like a bag, the pretrained model might generate images that include prior knowledge associations (e.g., a person carrying the bag) even without explicit prompts. In contrast, the fine-tuned model may lose this capability, leading to incoherent images.

\begin{figure}[ht]
  \centering
  \includegraphics[width=\linewidth]{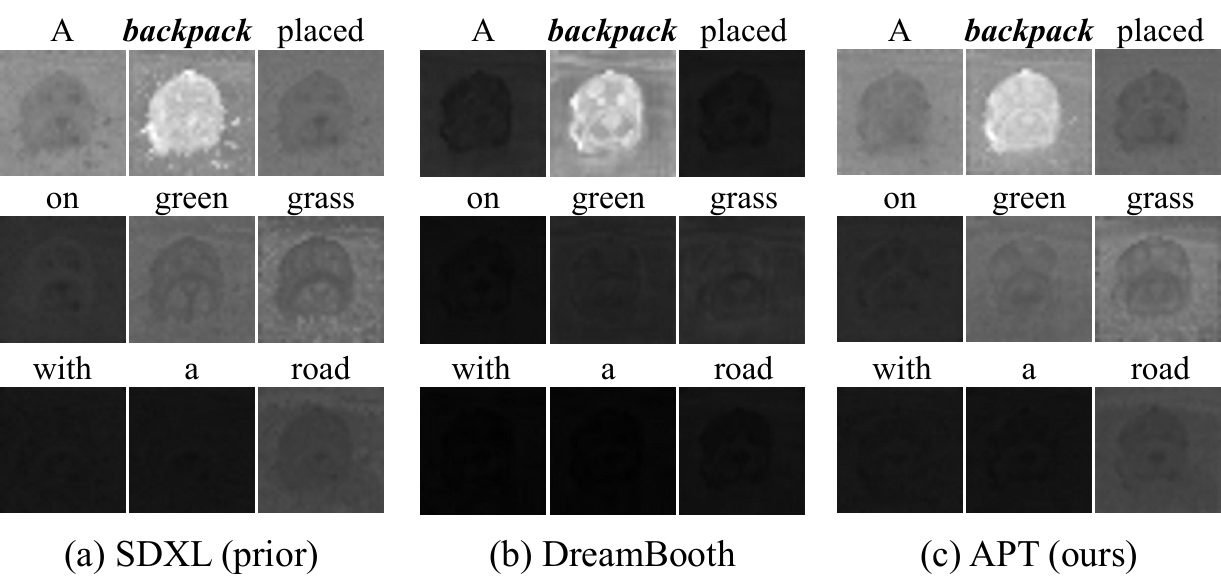}
  \caption{\textbf{Cross-Attention Map Comparison.} Visualization of cross-attention maps in text-conditioned image generation for (a) SDXL, (b) DreamBooth, and (c) APT. DreamBooth shows changes not only in the class token's map but also in overall attention maps, indicating shifts in how the model attends to different tokens after personalization.}
  \label{fig:attention}
  \vspace{-3mm}
\end{figure}


To address these issues, we introduce attention alignment for prior knowledge preservation, a regularization technique to align the cross-attention maps of the fine-tuned model with those of the pretrained model. Let $\mathbf{A}_{\theta,i}^{(l)}$ and $\mathbf{A}_{\phi,i}^{(l)}$ denote the cross-attention maps at layer $l$ and attention head $i$ for the fine-tuned model $\theta$ and the pretrained model $\phi$, respectively.
We define the attention regularization loss as:
\begin{equation}
\mathcal{L}_{\text{attn}} = \sum_{l}^{\text{layers}}\frac{1}{H}\left| \sum_{i=1}^{H}  \mathbf{A}_{\theta,i}^{(l)}(x_t; c^*, t) - \\  \sum_{i=1}^{H}  \mathbf{A}_{\phi,i}^{(l)}(x_t; c, t) \right|_2^2,
\label{eq:attn_reg}
\end{equation}
where $H$ is the number of attention heads. By differentiating between $c^*$ and $c$, we align the attention maps corresponding to the personalized concept with those of the general class, preserving prior knowledge.


By applying this regularization to \emph{all} text tokens, we ensure that the model maintains similar attention distributions across all tokens with the pretrained model. As training progresses, we observe that the influence of not only the identifier token but also other tokens changes, as shown in Figure~\ref{fig:attention}. By regularizing all tokens contributing to the representations, we aim to preserve the model's ability to understand the textual context, retaining the original semantic relationships and allowing the model to generate images that are coherent and contextually appropriate.

\subsection{Overall Training Objective}

The total training loss consists of the proposed regularization terms:
\begin{equation}
\mathcal{L}_{\text{total}} = \mathcal{L}_{\text{aptDM}} + \lambda_{\text{dist}} (\mathcal{L}_{\mu} + \mathcal{L}_{\sigma}) + \lambda_{\text{attn}} \mathcal{L}_{\text{attn}}
\end{equation}
where the hyperparameters $\lambda_{\text{dist}}$ and $\lambda_{\text{attn}}$ control the strength of the regularization terms.

In summary, our proposed method \ours~addresses the challenges of personalizing diffusion models with limited data by introducing adaptive training adjustments, representation stabilization, and attention alignment. By mitigating overfitting in an adaptive manner and preserving the statistical properties and attention distributions of the pretrained model, we enhance the ability to retain prior knowledge and maintain semantic coherence during fine-tuning.


%% file: sec/4_experiment.tex
\begin{figure*}[!ht]
  \centering
  \includegraphics[width=\linewidth]{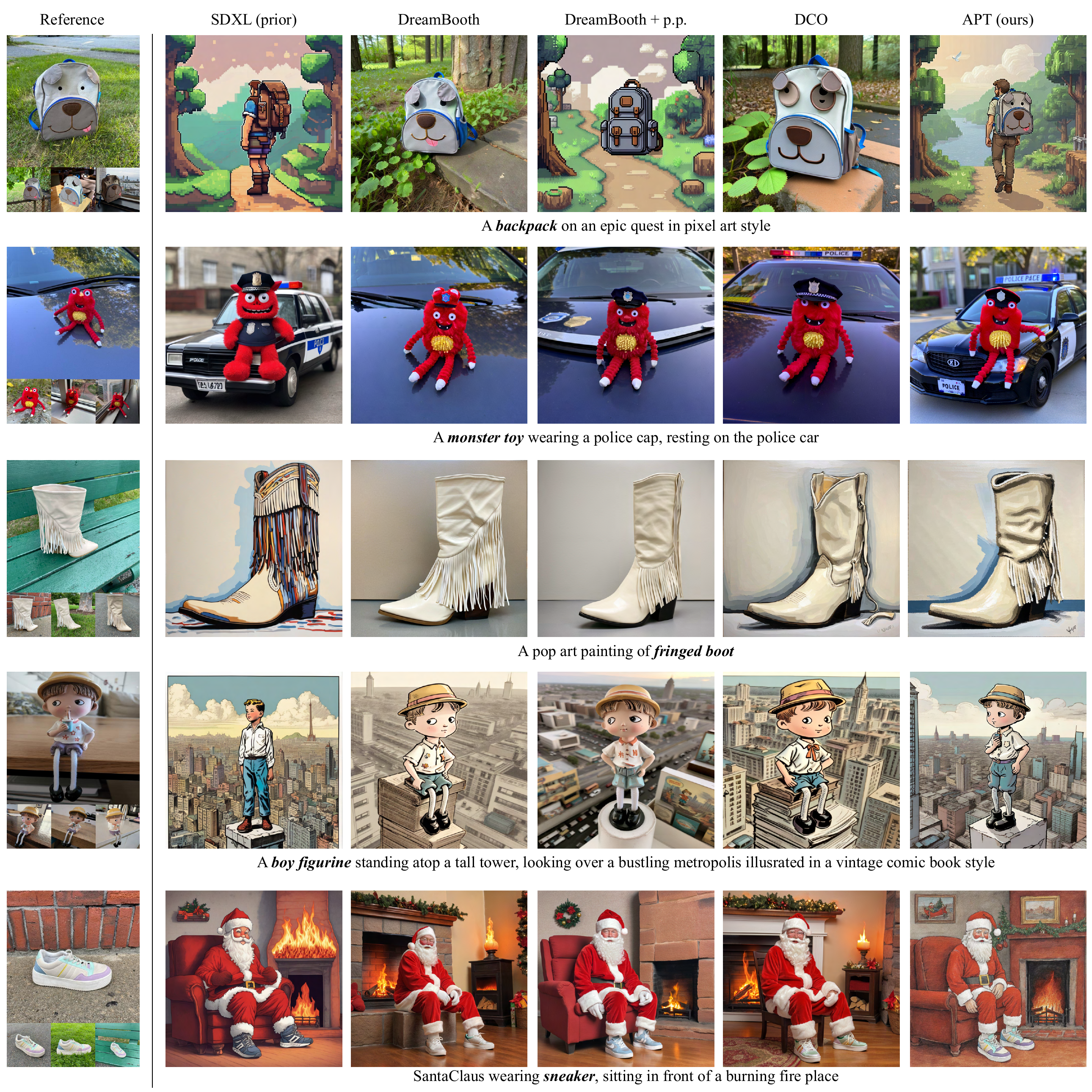}
  \caption{\textbf{Qualitative Comparison.} We present images generated by the pretrained model, DreamBooth, DreamBooth with Prior Preservation, DCO, and our method (\ours) across various data types and styles. Baseline methods tend to memorize textures and generate object-centric images, often lacking prior knowledge such as generating a person without explicit prompts. Objects are frequently zoomed in, with limited contextual and background details. In contrast, our method effectively integrates prior knowledge and generates images with better contextual alignment.
}
  \label{fig:main_fig}
\end{figure*}

\section{Experiments}
\label{sec:experiments}

In this section, we evaluate the effectiveness of our proposed APT in personalizing diffusion models with limited reference data. We compare APT with existing techniques through qualitative and quantitative comparisons, a user study, an ablation study, and an analysis of the overfitting indicator. Detailed ablation results and additional evaluations are provided in the Supplementary Material B.

\subsection{Experimental Setup}
\label{subsec:experimental_setup}


We adopt the pretrained Stable Diffusion XL model~\cite{sdxl}\footnote{Additional experiments using Stable Diffusion v2.1 are provided in Supplementary Material Section B.4.} as the foundation for all experiments. Our evaluations are conducted on commonly used datasets in personalization studies, specifically the \textbf{DreamBooth Dataset}~\cite{dreambooth} and the \textbf{Textual Inversion Dataset}~\cite{ti}. To generate captions for these images, we employ GPT-4o~\cite{gpt}, ensuring that the captions emphasize background descriptions while omitting explicit mentions of the target concept. This strategy prevents interference with the learning of the identifier and enables the model to focus on contextual details.

\vspace{-10pt}
\paragraph{Baselines}

We compare our APT method with the following baseline personalization techniques:

\begin{itemize}
\item \textbf{DreamBooth}: Combines DreamBooth~\cite{dreambooth} and Textual Inversion~\cite{ti} methods for concept learning.
\item \textbf{Custom Diffusion}~\cite{customdiffusion}: Performs efficient personalization by updating only the key and value of cross-attention.
\item \textbf{Direct Consistency Optimization (DCO)}~\cite{dco}: Addresses overfitting by regularizing the denoising process.
\end{itemize}

\vspace{-10pt}
\paragraph{Implementation Details}


Most methods, including ours, employ rank-32 LoRA~\cite{lora} for both U-Net and text encoder with a learning rate of $5 \times 10^{-5}$ and $5 \times 10^{-6}$ respectively, using a batch size of 1. Custom Diffusion~\cite{customdiffusion} does not use LoRA but instead fine-tunes the key and value of cross-attention, with a learning rate of $1 \times 10^{-5}$ and the same batch size of 1. The regularization weights are set to $\lambda_{dist} = 30$ and $\lambda_{\text{attn}} = 3 \times 10^{-4}$, and the maximum augmentation probability $p_{\text{max}}$ is 0.8. Further implementation details are provided in Supplementary Material A.


\subsection{Qualitative Analysis}
\label{subsec:qualitative_analysis}

Figure~\ref{fig:main_fig} shows a qualitative comparison between APT and baseline methods using identical text prompts. Our observations are summarized as follows:
\begin{itemize}
    \item \textbf{Scene Context and Background Preservation:} APT generates coherent backgrounds and naturally places objects (e.g., placing a backpack in a landscape), whereas baseline methods often generate overly zoomed-in views.
    \item \textbf{Prior Knowledge Preservation:} Unlike baselines that generate only the object, APT leverages the pretrained model’s prior knowledge to incorporate contextual elements such as human subject.
    \item \textbf{Textural and Stylistic Consistency:} APT replicates textures and styles from the pretrained model while maintaining semantic coherence.
    \item \textbf{Text Alignment:} APT faithfully follows textual instructions, achieving superior alignment with prompt details.
\end{itemize}

Overall, these qualitative results confirm that APT effectively achieves a balance between preserving contextual integrity and generating high-fidelity objects. Additional qualitative comparisons can be found in Supplementary material B.1.

\subsection{Quantitative Analysis}
\label{subsec:quantitative_analysis}

\input{table/main_table}

To quantitatively assess performance, we conduct a comprehensive evaluation measuring text-image similarity, image similarity, fidelity, and diversity across different methods using a diverse set of prompts from MS COCO~\cite{mscoco} captions. The results can be found in Table~\ref{tab:comprehensive_comparison}.

\begin{itemize}
\item \textbf{Text-Image Similarity}: We use the CLIP-T~\cite{CLIP} score and HPSv2~\cite{hpsv2} to evaluate how well generated images align with their corresponding text prompts. The CLIP-T score measures the cosine similarity between image and text embeddings, while HPSv2 assesses human preference for image-text alignment. Our method achieves superior text alignment on both metrics, particularly in HPSv2 scores, indicating better adherence to textual prompts while maintaining image quality.

\item \textbf{Image Similarity}: We compute the image similarity using DINOv2 features~\cite{dinov2}, which capture the semantic information of images. The similarities are calculated as the average pairwise cosine similarity between generated and reference images. Our method effectively preserves subject identity, outperforming SDXL while being comparable to or slightly lower than other baselines. This is due to our stronger emphasis on scene context over object-centric generation, which results in reduced zoomed-in artifacts in the generated images. Since DINOv2 similarity scores tend to favor closely cropped, object-centric images, our method's slightly lower similarity score reflects its ability to incorporate broader scene context rather than a deficiency in concept capture.

\item \textbf{Fidelity \& Diversity}: While precision and recall traditionally measure the fidelity and diversity of generated samples with respect to the real data distribution~\cite{assessinggen}, applying these metrics directly to diffusion model personalization is challenging. The few-shot nature of reference images prevents a reliable estimation of the real data distribution. Instead, we evaluate our method from a prior preservation perspective, measuring how well the personalized model maintains SDXL's generation capabilities. We establish two datasets: a source dataset generated using SDXL and a target dataset generated using personalized models. We then measure FID, Precision, and Recall between these source and target datasets. Our method outperforms other approaches, effectively preserving the original generation capabilities of SDXL while inheriting both its fidelity and diversity characteristics.
\end{itemize}

Overall, our method achieves competitive quantitative performance, validating its effectiveness in personalizing diffusion models with limited data.

\subsection{User Study}
\label{subsec:user_study}

\input{table/user_study}

We also conduct a user study to assess how well different models achieve personalization from a human alignment perspective. For simplicity in evaluation, we select DreamBooth and DCO as major baselines for comparison with our method. 20 participants blindly evaluate images from all three methods across 20 different prompts, with reference images and SDXL-generated images provided as prior knowledge on the following criteria (refer to Supplementary Material C for details):

\begin{itemize}
\item Assess the \textbf{text aligment} between the text prompt and the generated image, selecting the image that best reflects the detailed features of the text prompt.
\item Evaluate the \textbf{identity similarity} between objects in the training data and those in the generated images, along with the overall \textbf{image quality}.
\item Compare with images generated by the pretrained model, considering whether the generated images effectively preserve \textbf{prior knowledge} and are contextually appropriate.
\end{itemize}


\noindent As shown in Table~\ref{tab:comprehensive_comparison}, 56.1\% of the participants preferred the images generated by APT, compared to 22.8\% and 21.1\% for DreamBooth and DCO, respectively. This indicates that our method better aligns with the prompt and generates more visually appealing images than comparison methods.

\subsection{Ablation Study}
\label{subsec:ablation_study}

\begin{figure}[t]
  \centering
  \includegraphics[width=\linewidth]{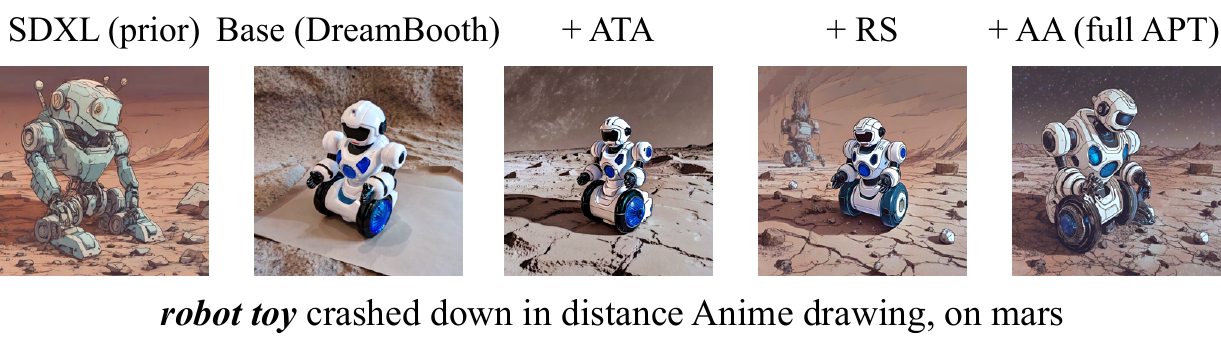}
  \caption{\textbf{Ablation Study of APT Components.} We evaluate the contribution of each component in our method by incrementally adding Adaptive Training Adjustment (ATA), Representation Stabilization (RS), and Attention Alignment (AA) to Base (DreamBooth).
  }
  \label{fig:ablation_study}
  \vspace{-3mm}
\end{figure}

To assess the contribution of each component in our method, we perform an ablation study by incrementally adding each component and observing the qualitative and quantitative effects (see Figure~\ref{fig:ablation_study} and Table~\ref{tab:comprehensive_comparison}).

\begin{enumerate}
\item \textbf{Base (DreamBooth)}: Base method without any of our proposed components
\item \textbf{+ATA}: Base + Adaptive Training Adjustment. With +ATA, zoomed-in artifacts are significantly reduced while improving Precision, Recall, and FID compared to Base, demonstrating the effectiveness of adaptive training strategies in mitigating overfitting.
\item \textbf{+RS}: Base + ATA + Representation Stabilization. Adding +RS further reduces texture memorization effects by reducing distribution shifts, significantly improving HPSv2 and the model's generalization ability while preserving the text fidelity of SDXL.
\item \textbf{+AA (full APT)}: Base + ATA + RS + Attention Alignment. With the full APT method (+AA), we observe better preservation of prior knowledge and improved metrics through regularization, allowing the personalized subject to be generated in a more coherent and contextually appropriate manner.
\end{enumerate}

As we add each component, we observe progressive improvements in image quality, text alignment, and preservation of prior knowledge. The full APT method generates the most coherent and contextually appropriate images. Additional examples and details are provided in Supplementary Material for further reference (see B.2).

\subsection{Analysis of Overfitting Indicator}
\label{subsec:overfitting_indicator_analysis}

\begin{figure}[!ht]
  \centering
  \includegraphics[width=\linewidth]{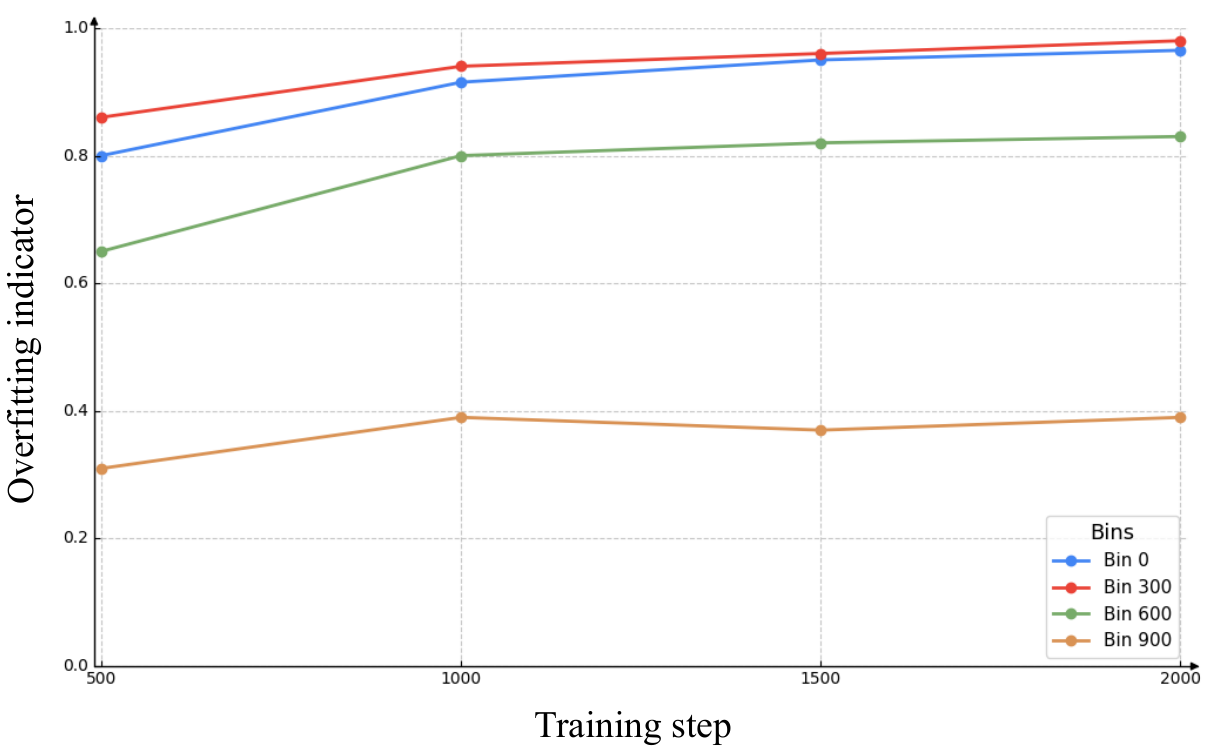}
  \caption{\textbf{Overfitting Indicator} The overfitting indicator $\gamma_t$ is plotted across training iterations for different time step bins.
  }
  \label{fig:gamma_plot}
  \vspace{-3mm}
\end{figure}

We analyze the behavior of the overfitting indicator $\gamma_t$ over training steps and time step bins to understand its influence on adaptive training adjustments. Figure~\ref{fig:gamma_plot} presents a plot of $\gamma_t$ across training iterations for different bins.

We observe that the overfitting indicator $\gamma_t$ increases more significantly for later time step bins (low noise levels) than for early timesteps (high noise levels). This indicates that overfitting occurs more rapidly at steps closer to the final image reconstruction, where the model begins to memorize specific details of the training data. The adaptive data augmentation and loss weighting respond accordingly, adjusting the training dynamics to mitigate overfitting where it is most pronounced. This adaptive mechanism helps maintain the stability of the denoising trajectory and preserves prior knowledge by dynamically adjusting to varying overfitting tendencies across different time steps.

%% file: table/main_table.tex
\begin{table*}[t]\centering
\label{tab:quant2} 
\begin{tabular}{lccccccc}
\toprule
\textbf{Method} & \multicolumn{2}{c}{\textbf{T-I Sim.}} & \multicolumn{1}{c}{\textbf{I Sim.}} & \multicolumn{3}{c}{\textbf{Fidelity\&Diversity}} & \multicolumn{1}{c}{\textbf{User Study (\%)}} \\\cmidrule(lr){2-3} \cmidrule(lr){4-4} \cmidrule(lr){5-7}
 & \textbf{CLIP-T}$\uparrow$ & \textbf{HPSv2}$\uparrow$ & \textbf{DINOv2}$\uparrow$ & \textbf{FID}$\downarrow$ & \textbf{Precision}$\uparrow$ & \textbf{Recall}$\uparrow$ \\\midrule
\midrule
SDXL (prior) & \textbf{0.666} & \textbf{0.295} & 0.625 & -- & -- & -- & -- \\\midrule
Custom Diffusion~\cite{customdiffusion} & 0.662 & 0.273 & 0.666 & 45.530 & 0.590 & 0.649 & -- \\
DCO~\cite{dco} & 0.662 & 0.277 & \textbf{0.687} & 52.298 & 0.548 & 0.660 & 21.1 \\\midrule
Base (Dreambooth)~\cite{dreambooth} & 0.661 & 0.272 & \underline{0.681} & 53.130 & 0.565 & 0.608 & \underline{22.8} \\
\ \ \textbf{+ATA}   & \underline{0.664} & 0.275 & 0.670 & 46.872 & 0.635 &0.680 & -- \\
\ \ \textbf{+RS} & \underline{0.664} & \underline{0.290} & 0.657 & \underline{42.663} & \textbf{0.701} &\underline{0.727} & -- \\
\ \ \textbf{+AA (full APT)} & \underline{0.664} & 0.288 & 0.660 & \textbf{41.967} & \underline{0.669} &\textbf{0.738} & \textbf{56.1} \\
\bottomrule
\end{tabular}
\caption{Quantitative comparison with baseline methods and ablation study of APT components. For evaluation, we use multiple metrics: Text-Image Similarity measured by CLIP-T and HPSv2 (higher values indicate better text alignment); Image Similarity measured by DINOv2 image-feature similarity (higher values indicate a closer resemblance to reference images); Fidelity\&Diversity measured by FID (lower is better) and Precision/Recall (higher is better); and User Study showing the percentage of participants selecting each method based on the criteria: preservation of prior knowledge, ability to capture the identity of reference images, and alignment with the prompt.}
\label{tab:comprehensive_comparison}
\end{table*}

%% file: table/user_study.tex

%% file: sec/5_conclusion.tex
\section{Limitations}
\label{sec:limitations}

While our proposed APT method effectively mitigates overfitting and preserves prior knowledge, it has certain limitations. The trade-off between preserving prior knowledge and learning new concepts is a fundamental challenge in text-to-image personalization. Although our work approaches Pareto-optimal solutions, some challenges remain. For example, when personalizing a ``monster toy'' intended to be cute, the strong prior associated with the word ``monster'' may cause the model to generate images with more monstrous appearances than desired. This issue arises because the identifier used during personalization is heavily influenced by the class word chosen for initialization. Adjusting the regularization weight $\lambda_{\text{attn}}$ associated with attention alignment can alleviate this problem by allowing more flexibility in how the model integrates prior knowledge. However, this introduces sensitivity to hyperparameters, which remains a limitation as it requires careful tuning for different concepts.

\section{Conclusion}
\label{subsec:conclusion}

We have presented APT, a novel method for personalizing diffusion models with limited data. By incorporating adaptive training adjustments, representation stabilization, and attention alignment, APT effectively mitigates overfitting and preserves prior knowledge. Our experiments demonstrate that APT outperforms existing methods, providing a robust solution for personalized generative modeling. Further research directions are discussed in Supplementary Material D.

%% file: sec/11_suppl.tex
\setcounter{section}{0}
\maketitlesupplementary
\renewcommand{\thesection}{\Alph{section}}

\section{Implementation Details}
\label{sec:appendix_implementation}

\paragraph{Additional Details}


All experiments are conducted using a single NVIDIA A100 GPU. For Representation Stabilization, we utilize the hidden states from the Upblocks of the U-Net at resolutions of $32 \times 32$ and $64 \times 64$. Additionally, for Attention Alignment, we employ the attention maps from the same Upblocks.
We use the AdamW optimizer~\cite{adamw} for training all models. The learning rate and other optimizer hyperparameters are set as described in the main text.
In Adaptive Data Augmentation, we apply zoom-out transformations with scales ranging from $1$ to $3$ and rotations within $\pm15$ degrees. We acknowledge that further experiments with additional augmentation types could be beneficial and are left for future work.
For the Exponential Moving Average (EMA) calculations, we set the smoothing factor $\alpha$ to $0.1$.
All generated images are generated using a Classifier-Free Guidance (CFG)~\cite{cfg} with scale of $7.5$. For DCO~\cite{dco}, to ensure a fair comparison, we use only CFG without Reward Guidance.

\paragraph{GPT-4o Caption Details}

Building upon Comprehensive Caption~\cite{dco}, we employ GPT-4o~\cite{gpt} to generate captions that emphasize on the background and context rather than the primary concept, allowing the token to learn the concept as directly as possible. We provide the reference data into GPT-4o and instruct it to describe each image, focusing on the surroundings and context while keeping the description of the central object as simple as possible.
We observe that when prompts contain detailed descriptions of the concept, the model struggles to learn those details effectively. By shifting the focus of captions to background and contextual elements, we ensure that the model learns rich and diverse information.
This approach not only enhances the learning of the desired concept through the token but also prevents the model from learning about non-target objects. By omitting detailed descriptions of the concept's color, texture, and other fine-grained details, we promote more robust learning and achieve better generalization when generating images conditioned on the learned concept.


\paragraph{Computations}
Our method requires an extra forward pass to retrieve the intermediate features of SDXL~\cite{sdxl}, which increases computational overhead—an approach also employed by the state-of-the-art method, DCO~\cite{dco}. However, since LoRA~\cite{lora} loaded into SDXL can be toggled on or off during the forward pass, our approach requires only the additional memory needed for the intermediate features, without the need to load a separate pretrained model.

\section{Additional Experimental Results}

\subsection{Qualitative Comparisons}
\label{sec:appendix_qualitative}

\begin{figure*}[!ht]
  \centering
  \includegraphics[width=\linewidth]{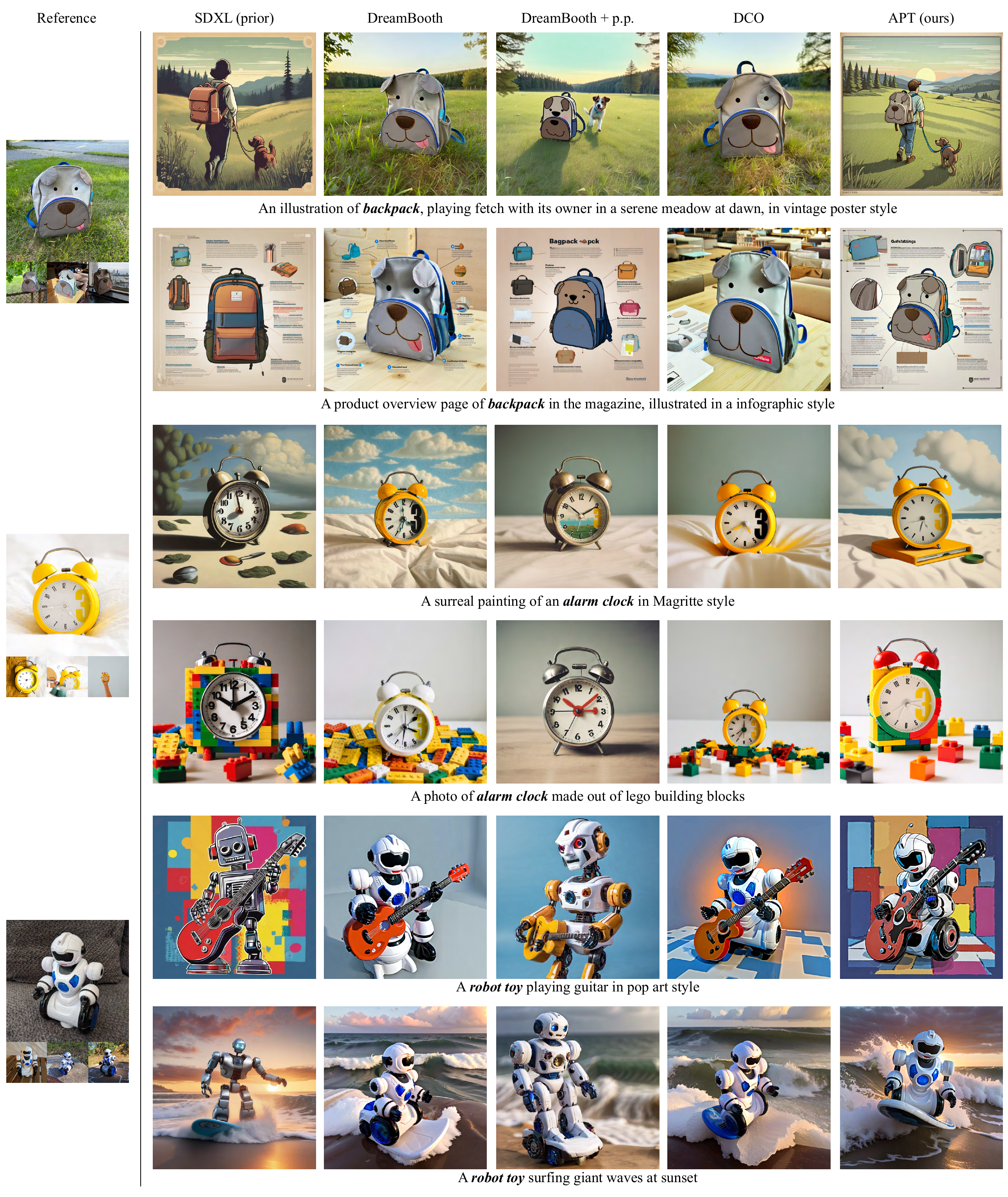}
  \caption{\textbf{Additional Qualitative Comparison.} We present four images generated by our method and two images from each of the baseline methods, including SDXL, DreamBooth \cite{dreambooth}, DreamBooth with prior preservation loss, and DCO \cite{dco}. Our method demonstrates superior performance in prior preservation, including text alignment, compared to these baselines.
}
  \label{fig:suppl_comparision1}
\end{figure*}

\begin{figure*}[!ht]
  \centering
  \includegraphics[width=\linewidth]{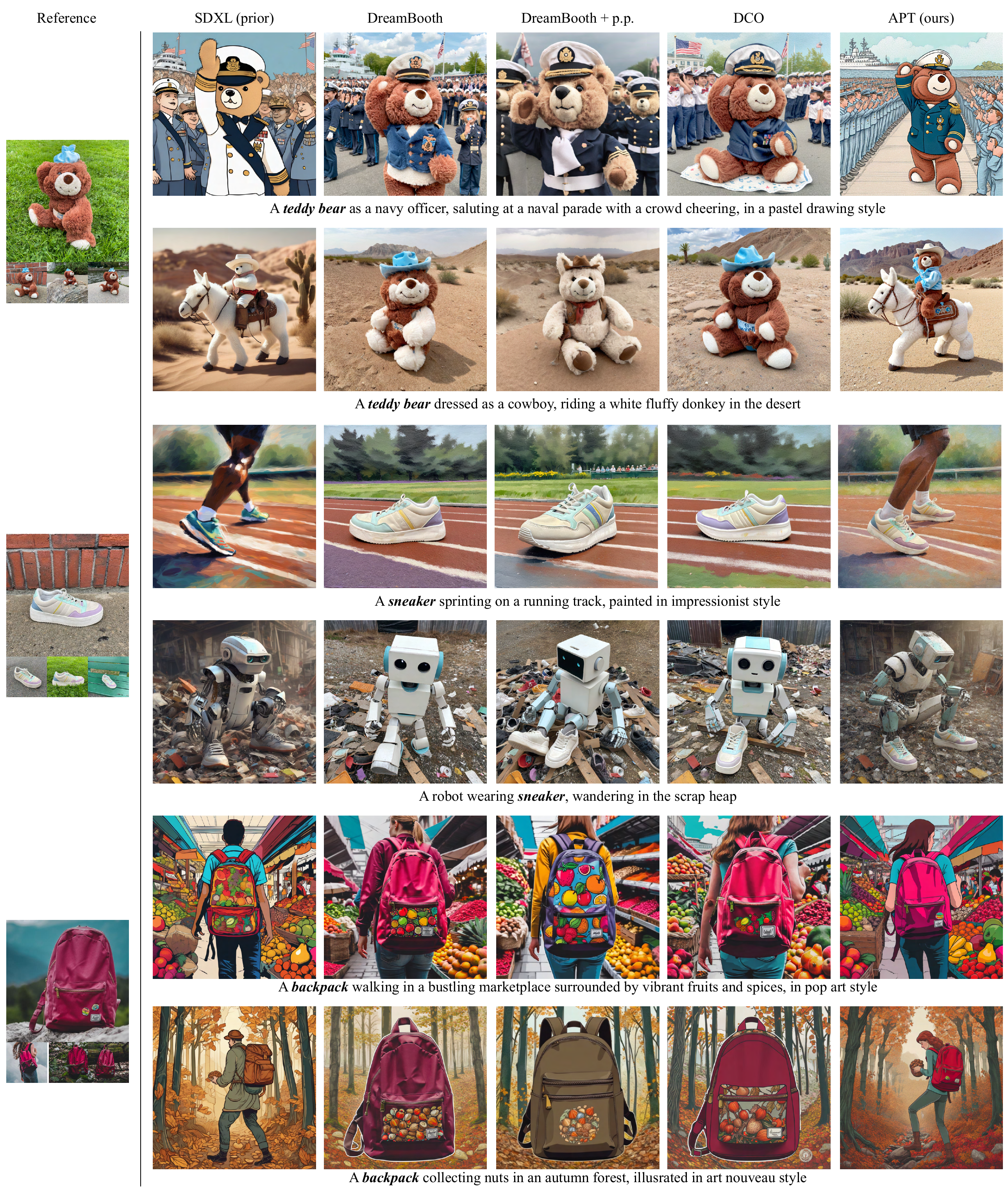}
  \caption{\textbf{Additional Qualitative Comparison.} We present four images generated by our method and two images from each of the baseline methods, including SDXL, DreamBooth \cite{dreambooth}, DreamBooth with prior preservation loss, and DCO \cite{dco}. Our method demonstrates superior performance in prior preservation, including text alignment, compared to these baselines.
}
  \label{fig:suppl_comparision2}
\end{figure*}

In Figure~\ref{fig:suppl_comparision1} and \ref{fig:suppl_comparision2}, we present additional qualitative comparisons between APT and baseline methods across diverse datasets and text prompts to demonstrate our model's superior performance. Our qualitative analysis reveals several key advantages of APT over existing approaches in four critical aspects described in Section~\ref{subsec:qualitative_analysis}. The baseline methods exhibit notable limitations in maintaining scene context and integrating prior knowledge, often generating overly focused, decontextualized images. For instance, when generating images of sneakers, baseline methods tend to generate isolated views that fail to capture the impressionist style specified in the prompt, while APT successfully incorporates these objects into coherent, prompt-aligned scenes that reflect the artistic direction. 

APT demonstrates remarkable capability in preserving prior knowledge from pretrained models, particularly in scenarios involving artistic style integration. When generating images of an alarm clock, APT successfully captures both the Magritte-style surrealist background and the distinctive texture of LEGO building blocks, while baseline methods struggle to maintain these artistic elements, often defaulting to conventional representations that lack the specified stylistic characteristics. This showcases the ability of APT to simultaneously handle multiple style requirements while maintaining object consistency.

\subsection{Ablation Study}
\label{sec:appendix_ablation}



\begin{figure}[!ht]
  \centering
  \includegraphics[width=\linewidth]{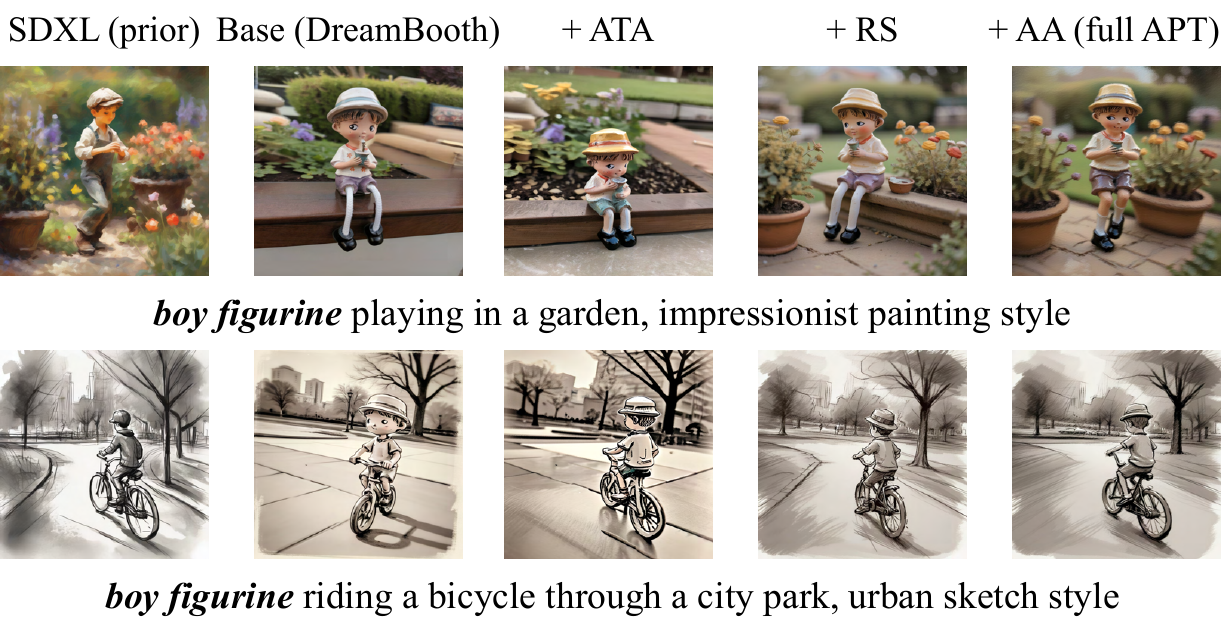}
  \caption{\textbf{Additional Ablation Study of APT Components.} We evaluate the contribution of each component in our method by incrementally adding Adaptive Training Adjustment (ATA), Representation Stabilization (RS), and Attention Alignment (AA) to Base (DreamBooth).
}
  \label{fig:suppl_ablation}
  \vspace{-3mm}
\end{figure}

We provide additional ablation results and analysis (see Table~\ref{tab:comprehensive_comparison} and Figure~\ref{fig:suppl_ablation}) to further demonstrate the impact of each component in our proposed APT framework. These results complement Section~\ref{subsec:ablation_study} and offer deeper insights into how each component contributes to mitigating overfitting and preserving prior knowledge.

\paragraph{Adaptive Training Adjustment (ATA)}
ATA immediately improves the baseline by mitigating overfitting. As shown in Table~\ref{tab:comprehensive_comparison}, applying ATA to the base model results in a modest increase in text-image similarity scores (with slight improvements in both CLIP-T and HPSv2) and a significant reduction in FID, which indicates better fidelity and diversity. Qualitatively, as illustrated in Figure~\ref{fig:suppl_ablation} (3\textsuperscript{rd} column), the ``zoomed-in'' effect observed in the base model's outputs is eliminated with ATA. The personalized object is no longer unnaturally enlarged or forced into the center; instead, it is rendered with greater flexibility in layout. This demonstrates that by introducing adaptive data augmentation and loss weighting, ATA effectively prevents the model from overfitting to a specific region or scale, thereby allowing for more natural object placement and pose variation.

\paragraph{Representation Stabilization (RS)}
Building on ATA, the addition of RS further improves the model’s performance. In Table~\ref{tab:comprehensive_comparison}, RS improves metrics related to prior preservation and alignment—for instance, increasing HPSv2 (indicating better prompt alignment) while slightly decreasing DINOv2 similarity (suggesting reduced over-tuning to reference details). Figure~\ref{fig:suppl_ablation} (4\textsuperscript{th} column) confirms that RS stabilizes intermediate representations during fine-tuning, which reduces the over-saturation of the subject's texture. By adjusting the distribution of latent features, RS prevents direct texture memorization, enabling the model to generalize better across different scenes and lighting conditions, while preserving the pretrained knowledge to adhere to the text prompt structure.

\paragraph{Attention Alignment (AA)}
Finally, incorporating AA (yielding the full APT model) unifies the benefits of the previous components and further refines the output. As shown in Table~\ref{tab:comprehensive_comparison}, AA helps the model maintain high text-image similarity while achieving low FID values. Supplementary metrics such as Recall also improve with AA, indicating enhanced output diversity. Figure~\ref{fig:suppl_ablation} (5\textsuperscript{th} column) demonstrates that AA improves semantic coherence: when applied, a personalized figurine is generated not only with its identity preserved but also with background elements and contextual cues that closely align with the prompt. AA achieves this by explicitly aligning the model's attention maps with those of the pretrained model, ensuring that attention is distributed across all prompt elements rather than being overly concentrated on the new concept token.

\paragraph{Overall Analysis}
The supplementary ablation study confirms that each component in APT contributes both individually and synergistically. ATA primarily mitigates spatial overfitting by freeing the object from a constrained, zoomed-in view. RS addresses feature-space overfitting by maintaining generalizable intermediate representations, and AA combats attention overfitting by ensuring a balanced focus across the entire prompt and scene. Although minor trade-offs (such as a slight decrease in precision with AA) are observed, they are more than compensated for by major gains in diversity and overall image coherence. Together, these results reinforce our claim that APT’s components are complementary and collectively enable state-of-the-art performance in personalized diffusion model training with limited data.




\subsection{Motivation for Adaptive Loss Weighting}
\label{sec:appendix_motivation}

Given a paired dataset of images $\mathbf{x}$ and captions $\mathbf{c}$, diffusion models are trained using a simplified version of the variational bound objective~\cite{ddpm, ldm}:
\begin{equation}
\mathcal{L}_{\text{simple}}(\theta; \mathcal{D}) := \mathbb{E}_{(\mathbf{x}, \mathbf{c}) \sim \mathcal{D}, \boldsymbol{\epsilon}, t} 
\left[ \omega(t) \|\boldsymbol{\epsilon} - \boldsymbol{\epsilon}_\theta(\mathbf{x}_t; \mathbf{c}, t) \|^2 \right],
\label{eq:ddpm}
\end{equation}
where $\mathbf{x}_t = \alpha_t\mathbf{x}_{t-1} + \sigma_t\boldsymbol{\epsilon}$ for $\boldsymbol{\epsilon} \sim \mathcal{N}(0, \mathbf{I}), t \sim \mathcal{U}(0,T)$. $\omega(t)$ is a weighting function allowing the model to focus on more challenging denoising tasks at larger timestep $t$ and make better sample quality. Min-SNR~\cite{min_snr} improves the convergence speed of training by considering the reverse process as a multi-task problem with varying difficulty levels and applying different clamped loss weights for each timestep interval. 

\begin{figure}[!ht]
  \centering
  \includegraphics[width=\linewidth]{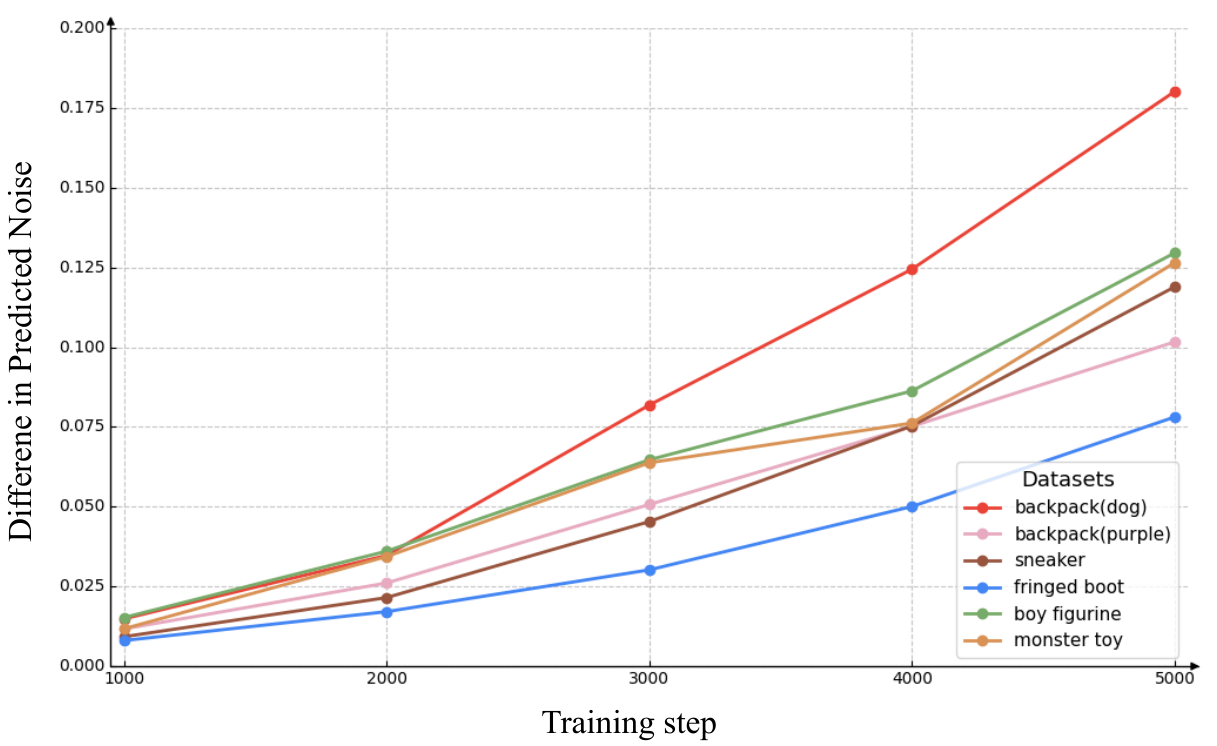}
  \caption{\textbf{Difference in Predicted Noise.} The difference in predicted noise between SDXL (prior) and DreamBooth~\cite{dreambooth} models is plotted over training iterations. Since the degree of overfitting varies across different datasets, we were motivated to detect overfitting during training and adjust the impact of the loss accordingly.
}
  \label{fig:suppl_indicator}
  \vspace{-3mm}
\end{figure}

However, since the training dynamics of personalizing diffusion models with limited data vary across different datasets, this necessitates excessive time and effort for hyperparameter optimization.
Figure~\ref{fig:suppl_indicator} illustrates the differences between the predicted noise of the pretrained SDXL model~\cite{sdxl} and that of the model fine-tuned using the DreamBooth~\cite{dreambooth} method, as follows:
\begin{equation}
\Delta{Noise}  =  \|\boldsymbol{\epsilon}_\phi(\mathbf{x}_t; \mathbf{c}, t)  - \boldsymbol{\epsilon}_\theta(\mathbf{x}_t; \mathbf{c}, t) \|^2
\label{eq:difference}
\end{equation}

As training progresses, the model loses the original distribution due to excessive shifts in the noise prediction, focusing solely on memorizing the training data and consequently degrading the model's ability to generalize to unseen prompts. This phenomenon appears similar across all datasets, but different overfitting patterns can be observed. At the end of training, the predicted noise difference between the model trained on the backpack (dog) dataset and the pretrained model is more than twice as large as that of the model trained on the fringed boot dataset. While severe overfitting may occur in specific datasets, this pattern does not generalize across all objects. Against this background, in Section~\ref{sec:ATA}, we introduce an Adaptive Overfitting Indicator that quantitatively measures the degree of overfitting during training in a dataset-dependent manner. Since the degree of overfitting varies across different datasets, our indicator adjusts adaptively during training. Additionally, we design a weighting scheme to reduce the impact of the loss accordingly when overfitting is detected, allowing the weights to vary based on the dataset rather than remaining fixed, as in previous approaches.

\subsection{Application to Stable Diffusion V2.1}
\label{sec:appendix_sd}

To demonstrate that our proposed APT is not only applicable to Stable Diffusion XL (SDXL) but also competitive when applied to other models, we conduct experiments using Stable Diffusion V2.1. Most existing personalization methods have been developed and evaluated on Stable Diffusion versions 1.4 or 2.1; thus, experimenting with V2.1 allows for a broader comparison with these methods.

\begin{figure*}[!ht]
  \centering
  \includegraphics[width=\linewidth]{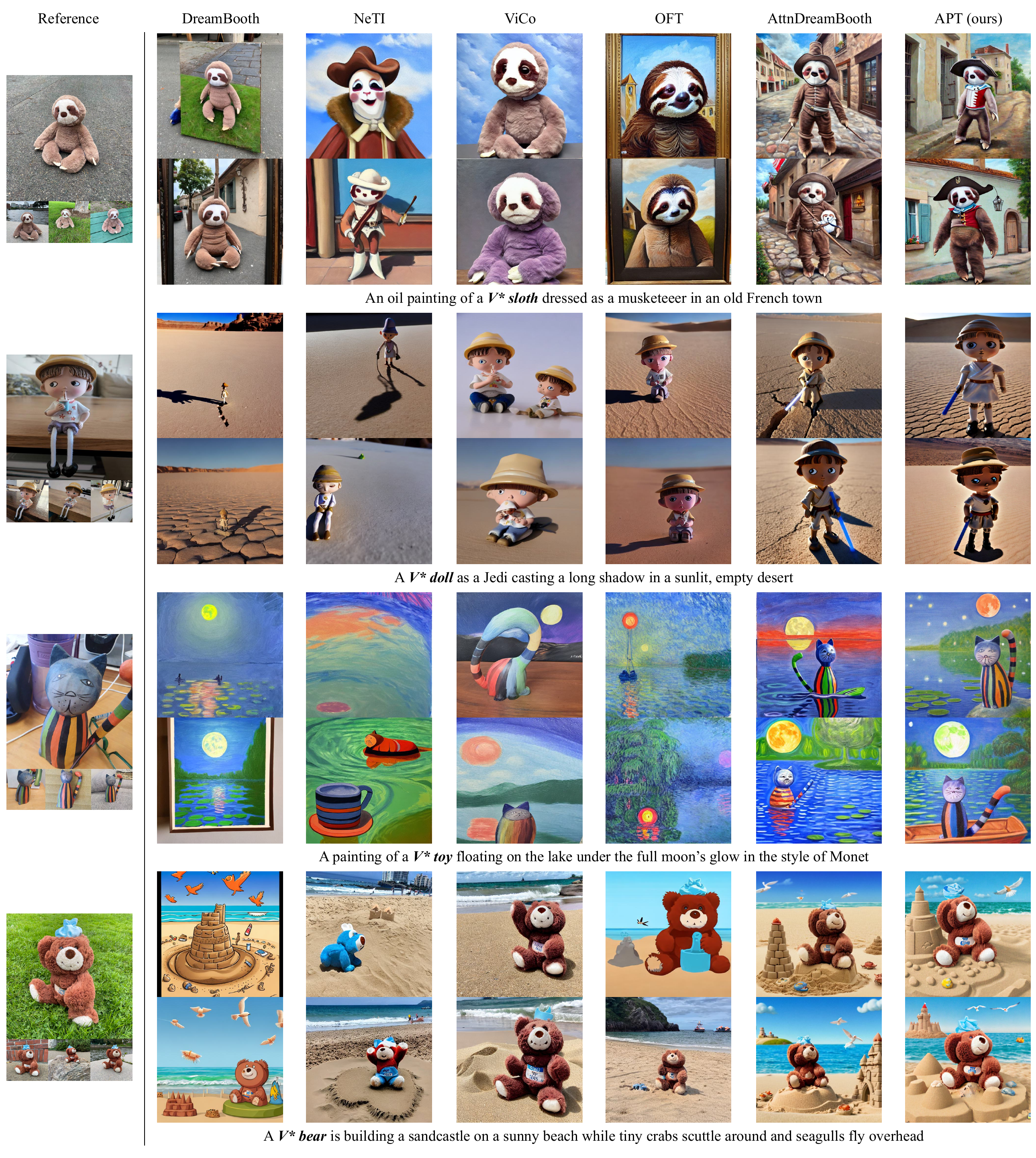}
  \caption{\textbf{Additional Qualitative Comparison on Stable Diffusion V2.1.} We compare \ours \ with other methods which are based on Stable Diffusion V2.1., including DreamBooth~\cite{dreambooth, ti}, NeTI~\cite{neti}, ViCo~\cite{vico}, OFT~\cite{oft}, and AttnDreamBooth~\cite{attndreambooth}. Two images from each of the baseline methods are collected from AttnDreamBooth~\cite{attndreambooth}. Our method outperforms baselines across various styles and contexts by effectively preserving prior knowledge.
}
  \label{fig:suppl_sd2.1}
\end{figure*}

In Figure~\ref{fig:suppl_sd2.1}, we compare APT with other methods based on Stable Diffusion V2.1, including DreamBooth~\cite{dreambooth,ti}, NeTI~\cite{neti}, ViCo~\cite{vico}, OFT~\cite{oft}, and AttnDreamBooth~\cite{attndreambooth}. All images except those generated by our method are directly taken from AttnDreamBooth~\cite{attndreambooth}.

For Stable Diffusion V2.1, we observe that the convergence speed of the overfitting indicator $\gamma$ differed from that in SDXL. Specifically, $\gamma$ converges more rapidly due to the characteristics of the model. To account for this, we adjust the calculation of $\gamma$ by using $T/10$ instead of $T$ in the exponential function, where $T$ is the total number of diffusion steps. All other hyperparameters are kept the same as in our experiments with SDXL.

We note that in models like Stable Diffusion V2.1, which have lower generation quality compared to SDXL, preserving prior knowledge can sometimes negatively affect the generated images. This is likely due to the limited capacity of the model to balance incorporating new concepts while maintaining existing knowledge. Despite this challenge, our method still outperforms the baselines across various styles and contexts by effectively preserving prior knowledge.

\section{User Study}
\label{sec:appendix_user_study}

In this section, we provide a detailed explanation of how the user study described in Section~\ref{subsec:user_study} is conducted.
Participants are presented with the following materials:

\begin{itemize}
\item \textbf{Reference Images}: The original images representing the target concept that the model was trained to learn.
\item \textbf{Prior Images}: Images generated by the pretrained model (SDXL) using the same noise seed and prompts without any personalization.
\item \textbf{Prompts}: The text descriptions used to generate images from the models.
\end{itemize}

Based on these materials, participants are asked to evaluate the generated images by considering the following aspects:

\begin{enumerate}
\item \textbf{Text Alignment}: Does the generated image align well with the given text prompt?
\item \textbf{Identity Preservation}: Is the generated image similar to the reference images?
\item \textbf{Prior Similarity}: Is the generated image similar to the composition of the prior image generated by the pretrained model?
\end{enumerate}

\begin{figure*}[!ht]
  \centering
  \includegraphics[width=\linewidth]{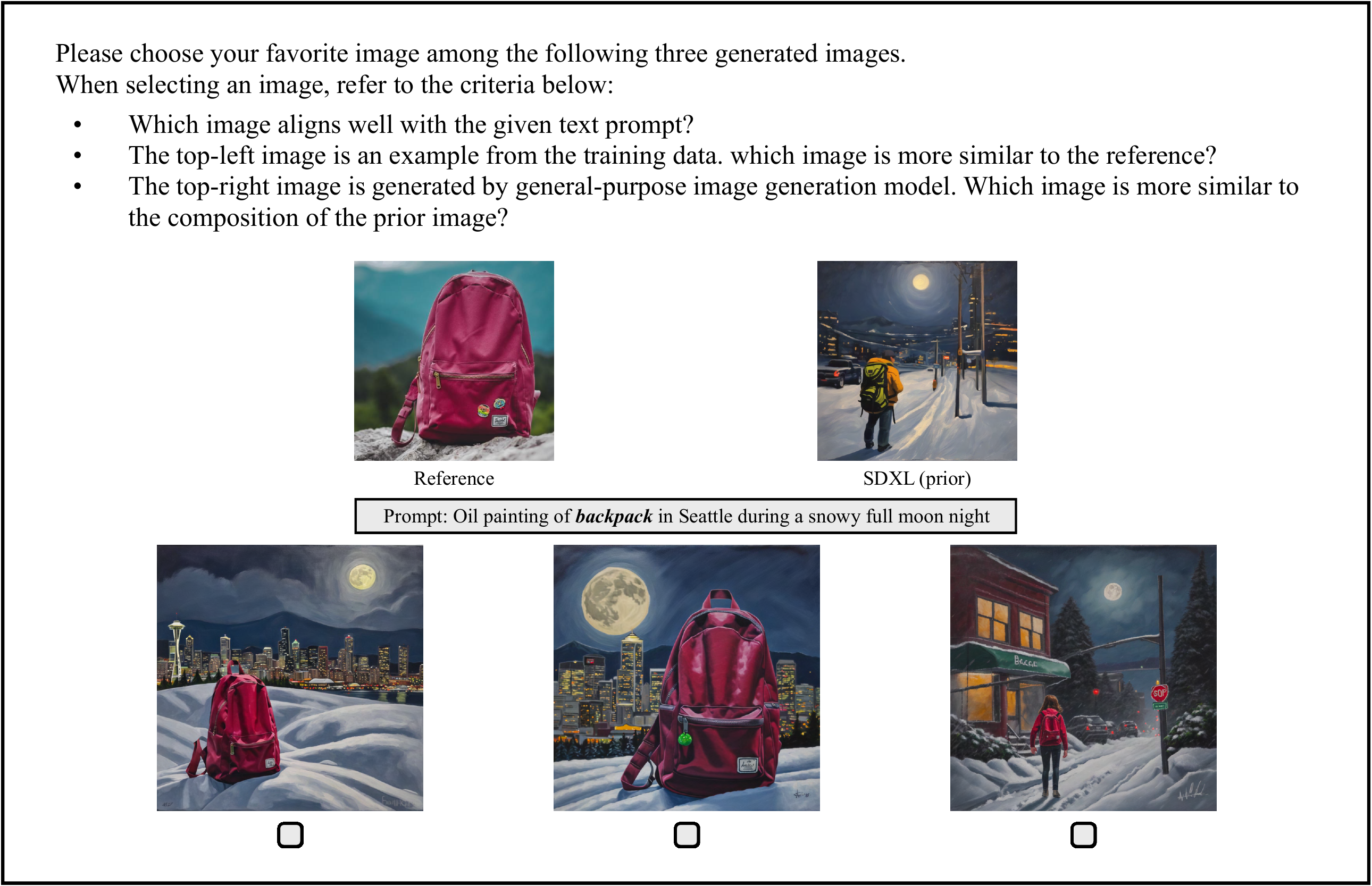}
  \caption{\textbf{User Study Example.} This shows the interface presented to users during the study.
}
  \label{fig:suppl_userstudy}
\end{figure*}

Participants are instructed to choose the image that best met all the criteria. Figure~\ref{fig:suppl_userstudy} shows the interface presented to users during the study. The results of the user study are summarized in Table~\ref{tab:comprehensive_comparison}.

\section{Future Work}
\label{sec:future_work}

In this section, we discuss potential areas for improvement and future research directions based on our observations.

\subsection{Reducing Memory and Computational Overhead}

Our method requires forwarding both the pretrained model $\phi$ and the fine-tuned model $\theta$ and comparing their attention maps and intermediate representations. This process requires more memory and computations, especially since attention maps from all layers are considered.

To address this issue, future work could focus on optimizing the computation by selecting only a subset of layers or resolutions for attention alignment and representation stabilization. For example, using attention maps and hidden states from specific layers or resolutions (e.g., only higher resolutions) that have the most impact on model performance could reduce computational load without significantly affecting the results.

\subsection{Combining Attention Alignment and Representation Stabilization}

Attention alignment and representation stabilization are closely related, as both aim to preserve the model's internal structures and prior knowledge. Given their close relationship, there is potential to combine these two components into a unified regularization term.

By formulating a joint regularization that considers both the attention maps and the hidden states simultaneously, we may achieve similar or improved performance with reduced computational complexity. Exploring this possibility could lead to a more efficient method that maintains the benefits of both components while mitigating computational overhead.

\afterpage{\clearpage}

\afterpage{\clearpage}

\afterpage{\clearpage}

